# Enhanced-alignment Measure for Binary Foreground Map Evaluation

**Deng-Ping Fan[1], Cheng Gong[1], Yang Cao[1], Bo Ren [1]*, Ming-Ming Cheng[1] and Ali Borji[2]**
[1] College of Computer and Control Engineering, Nankai University
[2] Center for Research in Computer Vision, Central Florida University
dengpingfan@mail.nankai.edu.cn, rb.nankai.edu.cn

## Abstract

The existing binary foreground map (FM) measures address various types of errors in either pixel-wise or structural ways. These measures consider pixel-level match or image-level information independently, while cognitive vision studies have shown that human vision is highly sensitive to both global information and local details in scenes. In this paper, we take a detailed look at current binary FM evaluation measures and propose a novel and effective **E-measure** (Enhanced-alignment measure). Our measure combines local pixel values with the image-level mean value in one term, jointly capturing image-level statistics and local pixel matching information. We demonstrate the superiority of our measure over the available measures on 4 popular datasets via 5 meta-measures, including ranking models for applications, demoting generic, random Gaussian noise maps, ground-truth switch, as well as human judgments. We find large improvements in almost all the meta-measures. For instance, in terms of application ranking, we observe improvement ranging from 9.08% to 19.65% compared with other popular measures.

## 1 Introduction

Please take a look at Fig. 1. You see the output of a binary foreground segmentation model and a random Gaussian noise map. While it is clear that the foreground map (FM) is much closer to the ground-truth (GT) map, to date the most common measures (*e.g.* , IOU [Everingham *et al.*, 2010], F1, and JI [Jaccard, 1901]) as well as recently proposed ones including Fbw [Margolin *et al.*, 2014] and VQ [Shi *et al.*, 2015] favor the noise map over the estimated map. This is one of the problems that we will address in this paper (see experiments Sec. 4.3). In order to solve it, we propose a novel measure that does much better than existing ones.

The comparison between a binary foreground estimated map and a human labeled ground-truth binary map is common in various computer vision tasks, such as image retrieval [Liu and Fan, 2013], image segmentation [Qin *et*

*Bo Ren is the corresponding author.

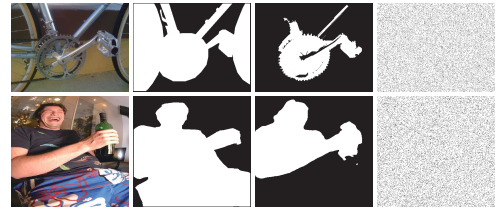

(a) Image   (b) GT   (c) FM   (d) Noise

Figure 1: Inaccuracy of current evaluation measures. A measure should score the FM (c) generated by a state-of-the-art algorithm a higher value than the random Gaussian noise map (d). Current common measures including IOU [Everingham *et al.*, 2010], F1/JI [Jaccard, 1901] , Fbw [Margolin *et al.*, 2014], CM [Movahedi and Elder, 2010], and VQ [Shi *et al.*, 2015] prefer the noise map. Only our measure correctly ranked (c) higher than (d).

*al.*, 2014], object detection, recognition [Kanan and Cottrell, 2010; Rutishauser *et al.*, 2004], foreground extraction [Blake *et al.*, 2004], and salient object detection [Hou *et al.*, 2018; Borji *et al.*, 2015], and is crucial for making statements regarding which models perform better.

Three widely used measures for comparing the foreground map (FM) and the GT include $F_\beta$ measure [Arbelaez *et al.*, 2011], Jaccard Index (JI) measure [Jaccard, 1901], and intersection over union (IOU) [Everingham *et al.*, 2010]. Various evaluations based on $F_\beta$-measures [Csurka *et al.*, 2013; Margolin *et al.*, 2014; Shi *et al.*, 2015] and other measures (*e.g.* [Movahedi and Elder, 2010; Villegas and Marichal, 2004; McGuinness and O'connor, 2010]) have been reported in the past. All of these evaluations, however, have used measures that address pixel-wise similarities and often discard structural similarities. Recently, Fan *et al.* [2017] proposed the structure measure (S-measure), which achieves great performance. However, this measure is designed for evaluating non-binary maps and some components (*e.g.* uniform distribution term) are not well defined for binary map case.

On the contrary, here, we propose a novel measure known as **E-measure** (Enhanced-alignment measure) which consists of a single term to account for both pixel and image level properties. We show that such an approach leads to an effective and efficient way for evaluating binary foreground maps as demonstrated in Fig. 2. Three foreground maps with colored borders (blue, red, yellow) are evaluated compared to the ground-truth map. Compared to 3 popular measures in-cluding Fbw, VQ, and CM, only our measure correctly ranks the three, considering both structural information and global shape coverage. We achieve this by taking into account image-level statistics (mean value of FM) and pixel-level matching jointly. Our measure (will be described in detail in Sec. 3) can correctly rank the estimated segmentation maps. Our main contributions are as follows:

- We propose a simple measure that consists of a compact term that simultaneously captures image level statistics and local pixel matching information. Using 5 meta-measures on 4 public datasets, we experimentally show that our measure is significantly better than traditional IOU, F1/JI, CM measures and the recently proposed ones including VQ, Fbw and S-measure.

- To assess the measures, we also propose a new meta-measure (SOTA vs. Noise) and build a new dataset. Our dataset contains 555 binary foreground maps which are ranked by humans. We use this dataset to examine the ranking consistency between current measures and human judgments.

| No. | Measure | Year | Pub | Pros | Cons |
|---|---|---|---|---|---|
| 1 | **IOU/F1/JI** [Jaccard, 1901] | 1901 | BSVSN | easy to calculate | losing image level statistics |
| 2 | **CM** [Movahedi and Elder, 2010] | 2010 | CVPRW | considering both region and contour | noise sensitive |
| 3 | **Fbw** [Margolin *et al.*, 2014] | 2014 | CVPR | assigning different weights for errors | error location sensitive, complicated |
| 4 | **VQ** [Shi *et al.*, 2015] | 2015 | TIP | weighting errors by psychological function | subjective measure |
| 5 | **S-measure** [Fan *et al.*, 2017] | 2017 | ICCV | considering structure similarity | focusing on non-binary map properties |

Table 1: Current evaluation measures summary.

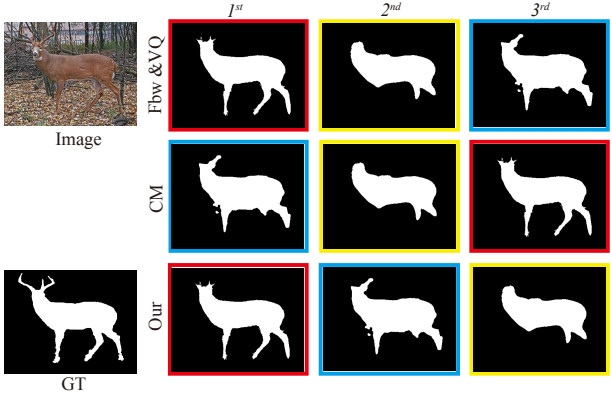

Figure 2: Demonstration of effectiveness of our measure. The ranking of binary foreground maps (after threshold) are generated by 3 state-of-the-art salient object detection models including DCL [Li and Yu, 2016], RFCN [Wang *et al.*, 2016] and DHS [Liu and Han, 2016]. All of the 3 different types of popular measures (CM, Fbw and VQ) fail to rank the maps correctly. However, our measure gives the right order.

## 2 Related Work

A summary of popular evaluation metrics for binary foreground map evaluation can be found in Tab. 1. Here, we explain these measures and discuss their pros and cons.

The $F_\beta$ measure [Arbelaez *et al.*, 2011; Cheng *et al.*, 2015; Liu *et al.*, 2011] is a common measure, which simultaneously

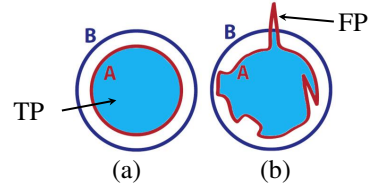

Figure 3: Limitations of region-based measures. The blue circle represents GT and red curve denotes FM. Based on IOU, F1/JI measures, the intersection in (b) is almost equal to the intersection in (a) when compared with GT circle (blue circle curve), although it has spikes, wiggles and shape differences [Movahedi and Elder, 2010].

considers $recall = \frac{TP}{TP+FN}$ & $precision = \frac{TP}{TP+FP}$:

$$F_\beta = \frac{(1+\beta^2)precision \cdot recall}{\beta^2 \cdot precision + recall}, \qquad (1)$$

where $\beta$ is a parameter to trade-off $recall$ and $precision$, True Positives (TP), True Negatives (TN), False Positives (FP), and False Negatives (FN) are 4 basic quantities. Setting $\beta=1$ leads to the classic $F1$ measure. Another widely used F1-based measure is the Jaccard Index (JI) [Jaccard, 1901], also known as the IOU measure:

$$JI = IOU = \frac{TP}{TP+FN+FP}. \qquad (2)$$

The $F1$ and IOU measures are related as: $JI = \frac{F1}{2-F1}$. Shi *et al.* [Shi *et al.*, 2015] proposed another measure for subjective object segmentation assessment. Essentially, their measure is also based on the F1 measure. Margolin *et al.* [2014] proposed a complicated measure called weighted $F_\beta$ measure (Fbw):

$$F_\beta^\omega = \frac{(1+\beta^2)Precision^\omega \cdot Recall^\omega}{\beta^2 \cdot Precision^\omega + Recall^\omega}. \qquad (3)$$

It assigns different weights to errors in different locations.

All the measures mentioned above are closely related with $F_\beta$. They can be estimated by considering each pixel position independently, and ignore important image level information, which leads to suboptimal performance in identifying noise (Fig. 1), structure errors (Fig. 2), and different shapes (Fig. 3).

Movahedi *et al.* [2010] proposed the contour mapping (CM) measure. This measure, however, is sensitive to noise (see Fig. 1), which results in poor performance, especially using meta-measure 3 as explained later (Sec. 4.3 & Tab. 2). A recently proposed measure known as S-measure [Fan *et al.*, 2017], focuses on non-binary foreground map (FM) evaluation. It considers the region-level structure similarity on a 2×2 grid over the segmentation map and the object-level properties (*e.g.*, uniform and contrast). However, these properties are not well defined for binary maps.

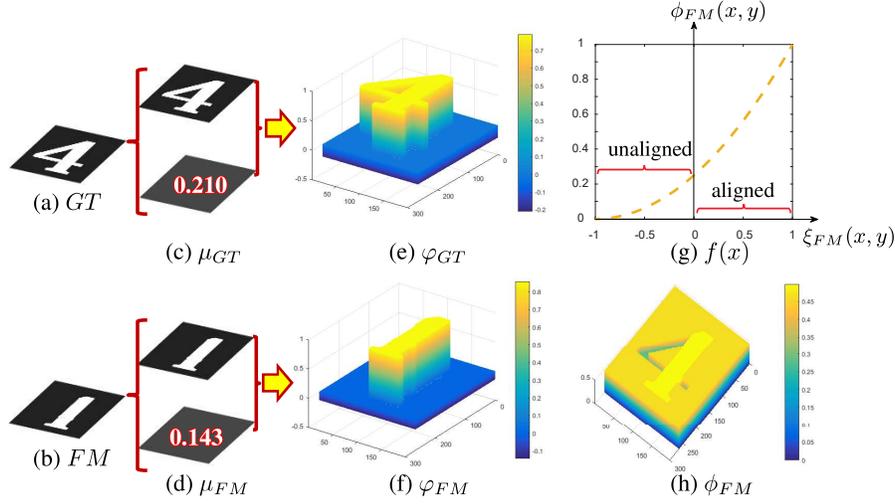

Figure 4: Our E-measure framework. (a) ground-truth map. (b) the estimate foreground map. (c) & (d) are the mean values map of GT & FM, respectively. (e) and (f) are the bias matrices calculated by (Eq. 4). (g) is the mapping function. (h) is the enhanced alignment matrix computed by (Eq. 6). *'aligned'* & *'unaligned'* donate those points which $GT(x,y) = FM(x,y)$ & $GT(x,y) \neq FM(x,y)$, respectively.

## 3 The Proposed Measure

In this section, we explain details of our new measure to evaluate binary foreground maps. An important advantage of our measure is its simplicity as it consists of a compact term that simultaneously captures ***global statistics*** and ***local pixel matching*** information. As a result, our measure performs better than the current popular measures. Our E-measure framework is demonstrated in Fig. 4.

### 3.1 Motivation

Despite the previous success of binary map measures, recent measures such as the S-measure are still not performing well enough on binary maps. It is often the case that these measures assign a higher score to a binary generic map than a state-of-the-art (SOTA) estimate segmentation map (see Sec. 4.2). The reason behind this is that in binary maps, S-measure puts emphasis on the luminance comparison, contrast comparison and the dispersion probability. However, while it makes sense to compute these terms for non-binary maps whose values are real numbers ranged in [0, 1] and treat the values as representing the probability that a pixel is owned by the foreground, in binary maps such properties are not well defined and less valid. As a result, using continuous assumptions can lead to erroneous evaluation on binary maps.

Cognitive vision studies have shown that human vision system is highly sensitive to structures (*e.g.* global information, local details) in scenes. Accordingly, it is important to consider local information and global information simultaneously when evaluating the similarity between an FM and a GT.

Based on the above observation, we design a novel measure that is tailored for binary map evaluation. Our measure works on well-defined properties of binary maps, and combines local pixel values and image-level mean value in one term, which helps to capture image-level statistics and local pixel matching information jointly. Experiments (Sec. 4) show that our measure performs better than other previous measures on binary maps.

### 3.2 Alignment Term

To design a compact term that simultaneously captures local pixel matching information and global statistics, we define a **bias matrix** $\varphi$ as the distance between each pixel-wise value of the input binary foreground map $I$ and its global mean $\mu_I$:

$$\varphi_I = I - \mu_I \cdot \mathbb{A}, \qquad (4)$$

where, $\mathbb{A}$ is a matrix in which all the element values are 1 and the size of $\mathbb{A}$ is identical to $I$. We compute two bias matrices $\varphi_{GT}$ and $\varphi_{FM}$ for the binary ground-truth map $GT$ and binary foreground map $FM$, respectively. $I \in \{GT, FM\}$. The bias matrix can be treated as the signal centering by removing the mean intensity from the signal. It can eliminate errors due to intrinsic variations, or large numerical differences.

Our bias matrix has strong relationship to the luminance contrast [Wang *et al.*, 2004]. Consequently, we consider the correlation (Hadamard product) between $\varphi_{GT}$ and $\varphi_{FM}$ as a simple and effective measure to quantify the bias matrix similarity. Therefore, we define an **alignment matrix** $\xi$ as following:

$$\xi_{FM} = \frac{2\varphi_{GT} \circ \varphi_{FM}}{\varphi_{GT} \circ \varphi_{GT} + \varphi_{FM} \circ \varphi_{FM}}, \qquad (5)$$

where, $\circ$ donates the Hadamard product. The alignment matrix $\xi_{FM}$ has the properties that $\xi_{FM}(x,y) \geqslant 0$ if and only if $\varphi_{GT}(x,y)$ and $\varphi_{FM}(x,y)$ have the same sign, *i.e.* the two inputs are *aligned* at this position $(x, y)$. The value of the alignment matrix element will depend on the global means, taking global statistic into account. These properties make Equ. (5) suitable for our purpose.

### 3.3 Enhanced Alignment Term

The absolute value of $\xi_{FM}(x, y)$ depends on the similarity of $\mu_{FM}$ and $\mu_{GT}$. When two maps are highly similar, further similarity between $\mu_{FM}$ and $\mu_{GT}$ may increase positive values at aligned positions and decrease negative values at unaligned positions. When summed together the total value of $\xi_{FM}(x, y)$ does not necessarily go up as we expected. Therefore, we need a mapping function that suppresses decrease (which means having smaller derivative value) at negative value ($\xi_{FM}(x, y) \leqslant 0$) regions and strengthens increase at positive value ($\xi_{FM}(x, y) \geqslant 0$) regions.

To achieve this goal, a "convex function" is needed. We have tested other forms of mapping functions such as higher-order polynomials or trigonometric functions, but have found the quadratic form ($f(x) = \frac{1}{4}(1+x)^2$ shown in Fig. 4 (g)) is a simple and effective function and works best in our experiments. Here, we use it to define the **enhanced alignment matrix** $\phi$ as:

$$\phi_{FM} = f(\xi_{FM}). \tag{6}$$

### 3.4 Enhanced Alignment Measure

Using the enhanced alignment matrix $\phi$ to capture the two properties (pixel-level matching and image-level statistics) of a binary map, we define our final **E-measure** as:

$$Q_{FM} = \frac{1}{w \times h} \sum_{x=1}^{w} \sum_{y=1}^{h} \phi_{FM}(x, y), \tag{7}$$

where $h$ and $w$ are the height and the width of the map, respectively. Using this measure to evaluate the foreground map (FM) and noise in Fig. 1, we can correctly rank the maps consistent with the application rank (See below).

## 4 Experiments

In this section, we compare our **E-measure** with 5 popular measures for binary foreground map evaluation on 4 public salient object detection datasets, as in [Fan *et al.*, 2017].

**Meta-Measure.** To test the quality of an evaluation measure, we use the meta-measure methodology. The basic idea is defining some desired criteria about the quality of the results and assessing how well a measure satisfies those criteria [Pont-Tuset and Marques, 2016]. We use 4 meta-measures proposed in [Margolin *et al.*, 2014; Pont-Tuset and Marques, 2016; Fan *et al.*, 2017], as well as a new one (Sec. 4.3) introduced here by us. Results are listed in Tab. 2.

**Datasets & Models.** The employed datasets include PASCAL-S [Li *et al.*, 2014], ECSSD [Xie *et al.*, 2013], HKU-IS [Li and Yu, 2015], and SOD [Martin *et al.*, 2001]. We use 10 state-of-the-art (SOTA) salient object detection models including 3 traditional ones (ST [Liu *et al.*, 2014], DRFI [Wang *et al.*, 2017], and DSR [Li *et al.*, 2013]) and 7 deep learning based ones (DCL [Li and Yu, 2016], RFCN [Wang *et al.*, 2016], MC [Zhao *et al.*, 2015], MDF [Li and Yu, 2015], DISC [Chen *et al.*, 2016], DHS [Liu and Han, 2016], and ELD [Lee *et al.*, 2016]) to generate non-binary maps. In order to further obtain foreground binary maps, we use the image-dependent adaptive thresholding method [Achanta *et al.*, 2009] to threshold non-binary maps.

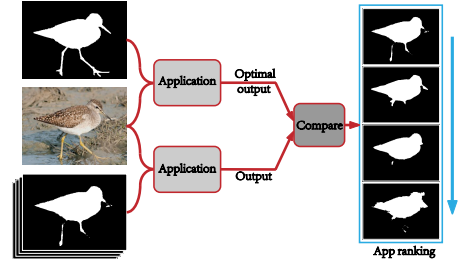

Figure 5: Application Ranking. To rank foreground maps according to an application, we compare the output when using the GT, to the output when using the FM foreground map. The more similar an FM foreground map is to a GT map, the closer its application's output should be to GT output.

### 4.1 Meta-Measure 1: Application Ranking

The first meta-measure specifies that the results of an evaluation measure to rank the foreground maps should be consistent with the results of application ranking. Fig. 5 illustrates the application ranking. Assume that the application's output when using GT maps is the optimal output. Then we feed a series of estimated maps to the application and obtain a sequence of outputs which have been ordered from the most similar to the most dissimilar. We compare the output sequence with the optimal output sequence. The more similar a map is to the GT map, the closer its application's output sequence should be to the GT output sequence.

As Margolin *et al.* [Margolin *et al.*, 2014] claimed, the applications including image retrieval, object detection and segmentation have similar results. For a fair comparison, we use the context-based image retrieval application as Margolin *et al.* to perform this meta-measure. Implementation of this application is mentioned in the Sec. **Application Realization** and other applications are implemented similarly.

Here, we use the $\theta = 1 - \rho$ [Best and Roberts, 1975] measure to examine the ranking correlation between measure ranking and application ranking. The value of $\theta$ falls in the range [0, 2]. Value of 0 means that the orders of measure ranking and application ranking are the same. Completely reverse orders give a value of 2.

In Tab. 2 we observe a significant improvement over the popular evaluation measures. Compared to the best prior measure, our measure improves the performance by 19.65%, 9.08%, 18.42% and 9.64% over PASCAL, ECSSD, SOD and HKU datasets, respectively. Fig. 2 illustrates an example of how well our measure predicts the preference of these applications.

**Application Realization.** Context-based image retrieval finds the most similar images to a query image in a dataset [Lew *et al.*, 2000]. The similarity is determined by various features such as color-histograms, color and edge directivity descriptor (CEDD). We use LIRE [Lew *et al.*, 2000] with CEDD to weigh the binary foreground maps.

Firstly, in order to ignore the background and get the foreground feature, we generate a combined GT image or FM image by combining the image with its GT map or FM map (see Fig. 6 (a)-(d)). The combined results are denoted by $GT_{combine} = \{G_1, \cdots, G_n\}$ and $FM_{combine} = \{F_1, \cdots, F_n\}$. Secondly, for each combined image we use LIRE to retrieve a list of 100 most similar combined images which are ordered from the most similar to the most dissimilar. The GT output, $GTout_i = \{G_{1-i}, \cdots, G_{100-i}\}$, is the ordered list returned when using the combined GT (*e.g.* $G_i$). The ordered score list is $GTscore_i = \{GTs_{1-i}, \cdots, GTs_{100-i}\}$. The score means the degree of similarity. Likewise, for the FM we obtain $Fout_i = \{F_{1-i}, \cdots, F_{100-i}\}$ and $Fscore_i = \{Fs_{1-i}, \cdots, Fs_{100-i}\}$. Thirdly, let $I_i = \{GTout_i \cap Fout_i\}$. We find $F_k$ which equals to $G_i$ in the $Fout_i$. If $F_k$ exists, indicating $G_i \in Fout_i$, we get the index $k$ as well as the corresponding score $Fs_{k-i}$. Each score $S_i$ of FM assigned:

$$S_i = \begin{cases} Fs_{k-i} + \frac{1}{k} + \frac{\|I_i\|}{100}, & G_i \in I_i \\ \frac{\|I_i\|}{100}, & otherwise \end{cases} \quad (8)$$

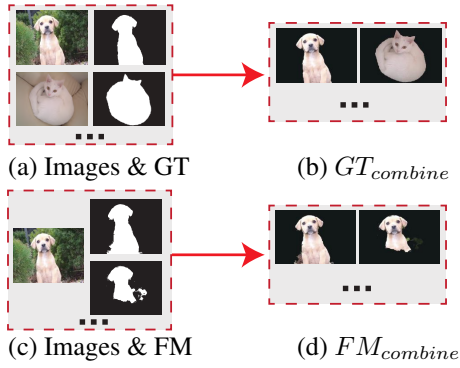

(a) Images & GT     (b) $GT_{combine}$

(c) Images & FM     (d) $FM_{combine}$

Figure 6: Combining the image with its foreground map.

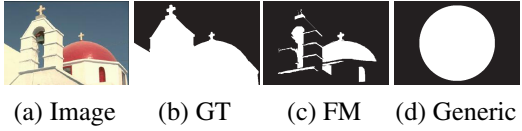

(a) Image    (b) GT    (c) FM    (d) Generic

Figure 7: Meta-measure 2: SOTA vs. Generic. An evaluation measure should give FM (c) generated by SOTA method a higher score than the generic map (d) that does not consider contents of the image.

### 4.2 Meta-Measure 2: SOTA vs. Generic Maps

The second meta-measure is that an evaluation measure should assign higher scores to maps obtained by the SOTA models than trivial maps without any meaningful contents. Here, we use a centered circle as the generic map. One example can be seen in Fig. 7. We expect that evaluating FM in (c) would generate a higher score than (d).

We counted the number of times a generic map scored higher than the mean score obtained by 10 SOTA models mentioned in Sec. 4 as mis-ranking rate. As suggested in [Margolin *et al.*, 2014], the mean score is robust to situations in which a certain model generates a poor result. The evaluation score of 10 maps should be higher than a threshold to choose as the "good map". Thus, about 80% good maps in the dataset have been selected to examine this meta-measure. The lower the mis-ranking rate is, the better the measure performs. Our measure outperforms the current measures over ECSSD, SOD and HKU-IS except on the PASCAL-S dataset.

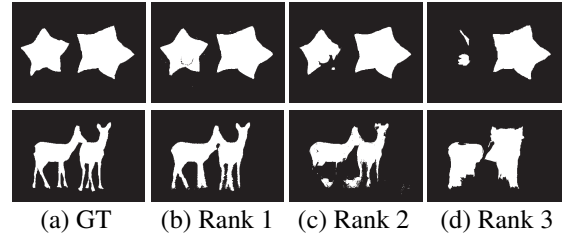

(a) GT    (b) Rank 1    (c) Rank 2    (d) Rank 3

Figure 8: Meta-measure 4: Human Ranking. Example images are from FMDatabase newly created by us.

### 4.3 Meta-Measure 3: SOTA vs. Random Noise

The property on which we based our third meta-measure is that an evaluation measure should prefer the map generated by a SOTA model over the random noise map on average.

We perform this experiment similar to the meta-measure 2 but this time we use the Gaussian random noise map instead of the generic map in Sec. 4.2. Our measure achieves the best performance since it considers both local pixel-matching and the global statistics jointly. It is to be noted that as stated in Sec. 4.2, the mean score is robust to single failure case of FM from a certain SOTA model. As a result the mean score on foreground maps from the group of SOTA models should always be higher than a score measured from noise map, however, only our measure and the S-measure achieve the lowest mis-ranking rate.

### 4.4 Meta-Measure 4: Human Ranking

The fourth meta-measure regards examining the ranking correlation between an evaluation measure and the human ranking. To the best of our knowledge, there is no such binary foreground map dataset ranked by human beings before. To create such a dataset, we randomly select the ranked maps by an application in meta-measure 1, from four above mentioned datasets including PASCAL, SOD, ECSSD and HKU. Then, we asked 10 subjects to rank these maps. We keep the maps for which all of the viewers agree in their rankings. We name our dataset **FMDatabase**[1] which contains 185 images. Each of the images comes with 3 ranked estimated maps (555 maps in total).

To give a quantitative assessment of the correlation between human ranking and measure ranking, we also use the $\theta$ measure (mentioned in meta-measure 1) to examine this meta-measure. The lower the score is, the more consistent an evaluation measure is for human ranking. As it can be seen, our measure outperforms other measures. Fig. 2 illustrates an example of how well our measure predicts the preference using human ranking.

### 4.5 Meta-Measure 5: Ground Truth Switch

The property on which we base the fifth meta-measure is that the score of a "good map" should decrease when we use a wrong GT map. We analyzed 4 popular datasets (PASCAL, SOD, ECSSD, HKU) and found that a map is considered as "good" when it scores (using F1 measure) at least 0.8 out of 1. We follow Margolin *et al.* [Margolin *et al.*, 2014] to calculate this meta-measure. We count the percentage of times that an evaluation measure assigns a higher score when using the wrong GT map. We found that all the evaluation measures perform well (The average result on 4 datasets are: VQ with 0.000925%, CM with 0.001675%, IOU/JI/F1 with 0.00515%, S-measure with 0.0014% and ours with 0.0523%). Our measure has a 0.05% gap relative to other measures.

---
[1]FMDatabase: http://dpfan.net/e-measure/

| Measure | PASCAL | | | ECSSD | | | SOD | | | HKU | | | |
|---|---|---|---|---|---|---|---|---|---|---|---|---|---|
| | MM1 | MM2 | MM3 | MM1 | MM2 | MM3 | MM1 | MM2 | MM3 | MM1 | MM2 | MM3 | MM4 |
| **CM** | 0.610 | 49.78% | 100.0% | 0.504 | 34.62% | 100.0% | 0.723 | 29.89% | 56.22% | 0.613 | 25.26% | 100.0% | 1.492 |
| **VQ** | 0.339 | 17.97% | 15.32% | 0.294 | 7.445% | 6.162% | 0.335 | 9.143% | 14.05% | 0.331 | 3.067% | 1.800% | 0.161 |
| **IOU/F1/JI** | 0.307 | 9.426% | 5.597% | 0.272 | 4.097% | 1.921% | 0.342 | 4.571% | 6.857% | 0.303 | 0.900% | 0.197% | 0.124 |
| **Fbw** | 0.308 | 5.147% | 4.265% | 0.280 | 2.945% | 1.152% | 0.361 | 6.286% | 5.714% | 0.312 | 0.535% | 0.083% | 0.149 |
| **S-measure** | 0.315 | 2.353% | 0.000% | 0.279 | 1.152% | 0.000% | 0.374 | 1.714% | 0.000% | 0.312 | 0.141% | 0.000% | 0.140 |
| **Ours** | **0.247** | 3.093% | **0%** | **0.247** | **0.641%** | **0%** | **0.273** | **0.571%** | **0%** | **0.274** | **0.084%** | **0%** | **0.121** |

Table 2: Quantitative comparison between the E-measure and current measures on 4 meta-measures. The best result is highlighted in **bold.** MM: Meta-Measure. These differences are all statistically significant at the $\alpha < 0.05$ level.

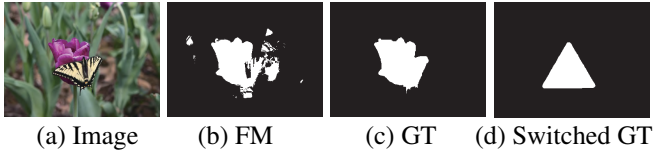

(a) Image    (b) FM    (c) GT    (d) Switched GT

Figure 9: Meta-measure 5: Ground-truth switch. An evaluation measure should assign the "good" map (b) a higher score when using the right ground-truth (c) as the reference than using the randomly switched ground-truth (d).

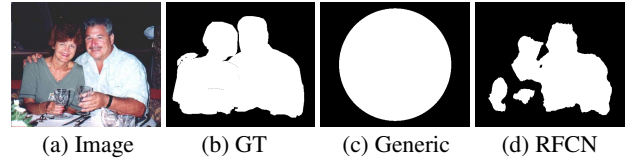

(a) Image    (b) GT    (c) Generic    (d) RFCN

Figure 10: Failure case of our E-measure in MM2. Due to ignore the semantic information, our E-measure ranked (c) higher than (d).

## 5 Conclusion and Future Work

In this paper, we analyzed various binary foreground evaluation measures that consider errors in different levels including pixel, region, boundary and object level. They can be classified as considering either pixel-level errors or image-level errors independently. To solve this shortcoming, here we proposed the simple **E-measure** which simultaneously considers both types of errors. Our measure is highly effective and efficient. Extensive experiments using 5 meta-measures demonstrate the effectiveness of our measure compared to existing measures on 4 popular datasets. Finally, we created a new dataset (740 maps) which consists of 185 ground-truth maps and 555 human ranked maps to examine the correlation between evaluation measures and human judgments.

**Limitation.** Compared to our metric, S-measure is mainly designed for tackling structural similarity. Images in PASCAL dataset have more structural objects than the other 3 datasets (ECSSD, SOD, HKU-IS). Therefore, S-measure is slightly better than our metric in PASCAL dataset. One failure case can be found in Fig. 10.

**Future Work.** We will investigate the potential to propose a new segmentation model based on the E-measure in our future work. Besides, our metric consists of simple derivable functions, so a new loss function based on the E-measure can be developed. To help future explorations in this area, our code and dataset will be made publicly available on the web.

## Acknowledgments

This research was supported by NSFC (NO. 61620106008, 61572264), Huawei Innovation Research Program, and Fundamental Research Funds for the Central Universities.